\documentclass[runningheads]{llncs}

\usepackage{bbding}

\usepackage{marvosym}
\usepackage{float}
\usepackage{amsmath}
\usepackage{dutchcal}
\usepackage{hyperref}
\usepackage{subfigure}
\usepackage{graphicx}
\usepackage{color}
\usepackage{amssymb}
\usepackage{booktabs}
\usepackage{makecell}
\usepackage{svg}
\usepackage{datetime}
\usepackage{multirow}
\usepackage{cases}
\usepackage{textcomp}

\newcommand{\figref}[1]{Fig.~\ref{#1}}

\begin{document}
\title{TT-DF: A Large-Scale Diffusion-Based Dataset and Benchmark for Human Body Forgery Detection}
\titlerunning{TT-DF: A Dataset and Benchmark for Body Forgery Detection}

\author{Wenkui Yang\inst{1,2} \and Zhida Zhang\inst{1,2} \and Xiaoqiang Zhou\inst{1,3} \and Junxian Duan\inst{1} \and Jie Cao\inst{1}$^{(\textrm{\Letter})}$}
\authorrunning{W. Yang et al.}
\institute{MAIS \& NLPR, Institute of Automation, Chinese Academy of Sciences, Beijing, China \and School of Artificial Intelligence, University of Chinese Academy of Sciences, Beijing, China \and University of Science and Technology of China, Hefei, China \\ \email{yangwenkui20@mails.ucas.ac.cn}, \email{\{zhida.zhang,jie.cao\}@cripac.ia.ac.cn},  \email{xq525@mail.ustc.edu.cn}, \email{junxian.duan@ia.ac.cn}}


\maketitle              
\begin{abstract}
The emergence and popularity of facial deepfake methods spur the vigorous development of deepfake datasets and facial forgery detection, which to some extent alleviates the security concerns about facial-related artificial intelligence technologies. 
However, when it comes to human body forgery, there has been a persistent lack of datasets and detection methods, due to the later inception and complexity of human body generation methods. 
To mitigate this issue, we introduce \textbf{T}ik\textbf{T}ok-\textbf{D}eep\textbf{F}ake (TT-DF), a novel large-scale diffusion-based dataset containing 6,120 forged videos with 1,378,857 synthetic frames, specifically tailored for body forgery detection.
TT-DF offers a wide variety of forgery methods, involving multiple advanced human image animation models utilized for manipulation, two generative configurations based on the disentanglement of identity and pose information, as well as different compressed versions.
The aim is to simulate any potential unseen forged data in the wild as comprehensively as possible, and we also furnish a benchmark on TT-DF.
Additionally, we propose an adapted body forgery detection model, \textbf{T}emporal \textbf{O}ptical \textbf{F}low \textbf{Net}work (TOF-Net), which exploits the spatiotemporal inconsistencies and optical flow distribution differences between natural data and forged data. 
Our experiments demonstrate that TOF-Net achieves favorable performance on TT-DF, outperforming current state-of-the-art extendable facial forgery detection models. For our TT-DF dataset, please refer to \href{https://github.com/HashTAG00002/TT-DF}{\textit{github.com/HashTAG00002/TT-DF}}.


\end{abstract}
\section{Introduction}
With the rise of Generative Adversarial Networks (GANs) \cite{goodfellow2014generative} and diffusion models \cite{sohl2015deep}, generative methods have seen significant development in recent years, making many applications accessible even to amateur users \cite{cao2022scoremix,zhou2021image,zhou2024ristra}.
These generative models have the potential for malicious use, such as illegal commercial, pornographic, or fraudulent activities. 
Hence, there arises a pressing necessity for forgery detection models capable of discerning synthetic data to alleviate the negative impacts they may entail.

Across various types of forgery, facial forgery has long been a significant focus for both researchers and casual users, with numerous facial forgery datasets \cite{rossler2019faceforensics++,li2020celeb,DFDC,zhou2021face} and detection models correspondingly proposed. 
However, the field of body forgery detection has not received equivalent attention so far, despite the recent discovery of image animation models \cite{karras2023dreampose,wang2023disco,chang2023magicdance,xu2023magicanimate,hu2023animate} by the general public for their potential in malicious forgery targeting human bodies.
Currently, visually convincing synthetic videos can often be achieved with careful manual selection. This further highlights the importance of dedicated datasets and specialized detection models for body forgery, which are still lacking in this field.

\begin{figure} 
\centering
\includegraphics[width=0.9\textwidth]{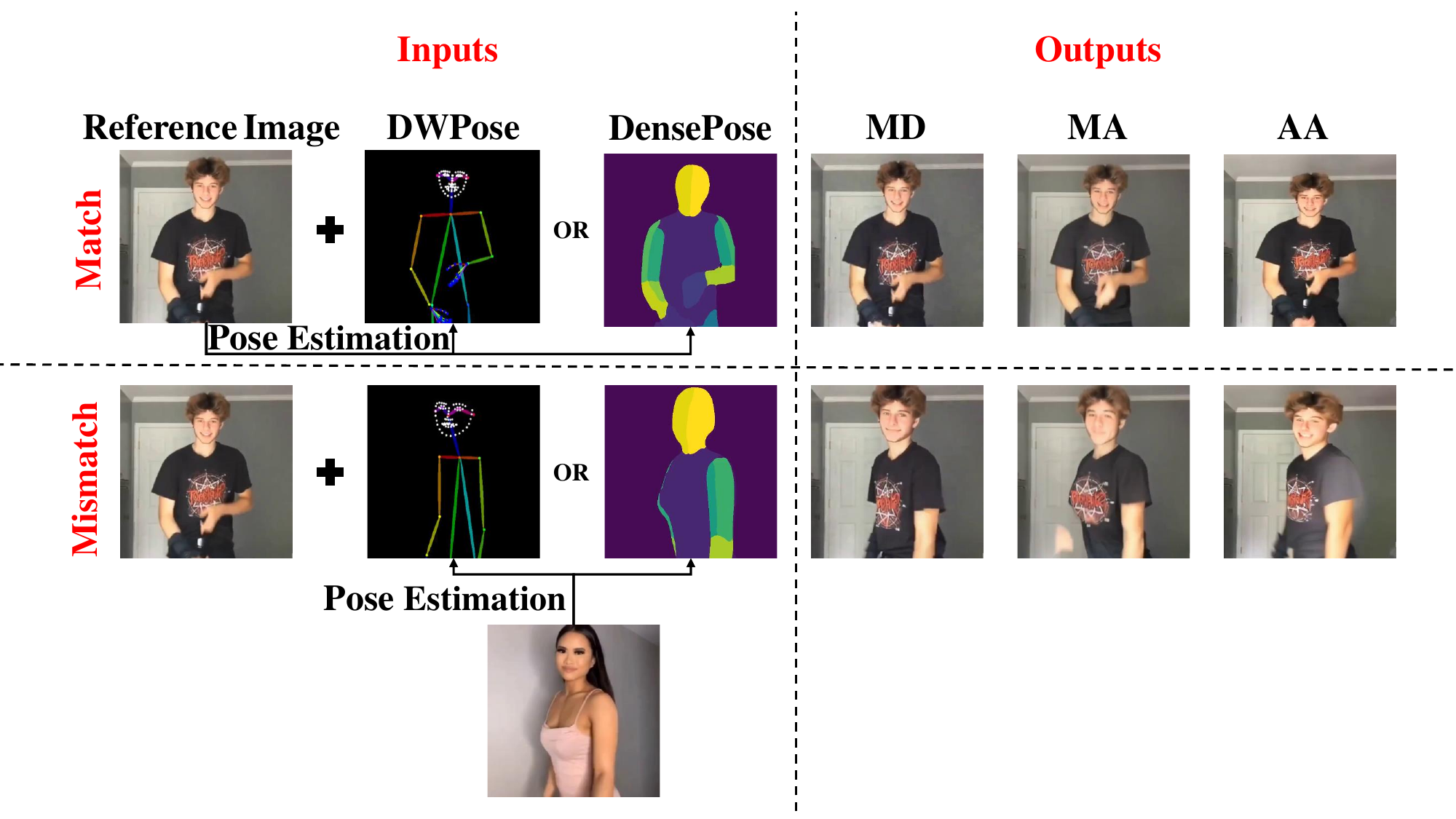}
\caption{Overview diagram of our proposed \textbf{T}ik\textbf{T}ok-\textbf{D}eep\textbf{F}ake (TT-DF) dataset. TT-DF employs three animation methods based on latent diffusion models: MagicDance (MD) \cite{chang2023magicdance}, MagicAnimate (MA) \cite{xu2023magicanimate}, and AnimateAnyone (AA)\cite{hu2023animate}, disentangling pose sequences from identity information. Two subsets, $Match$ and $Mismatch$, are generated according to whether the pose sequences and reference images match.} 
\label{fig1}
\end{figure}

In this work, we generate a novel dataset, \textbf{T}ik\textbf{T}ok-\textbf{D}eep\textbf{F}ake (TT-DF), for the development of human body forgery detection. 
The main advantages of TT-DF lie in the following aspects: 
(1) \textbf{Pioneer.} 
To the best of our knowledge, despite the longstanding presence of human image animation techniques, the corresponding concept of body forgery detection has not been explicitly established in the forgery detection field. Hence, TT-DF represents a pioneering initiative dedicated to this emerging body forgery area.
(2) \textbf{Large-scale.} 
TT-DF includes 6,120 forgery videos in total, comprised of 1,378,857 synthetic frames, involving distinct generative models and configurations. 
We also apply H.264 compression to the raw dataset and generate two compressed versions.
(3) \textbf{Multiple forgery methods.} 
TT-DF follows the latest body image animation works based on latent diffusion models \cite{rombach2022high}, selecting three advanced generation models for manipulation: MagicDance \cite{chang2023magicdance}, MagicAnimate \cite{xu2023magicanimate}, and AnimateAnyone \cite{hu2023animate}. All of these methods need two relative modalities as inputs: pose and identity information. 
Accordingly, for each manipulation model, we adopt two generative configurations and obtain two subsets, $Match$ and $Mismatch$, based on whether the ID information and the pose sequence are extracted from the same real video, as shown in \figref{fig1}. In addition, to provide a benchmark for the proposed TT-DF, we implement and evaluate several classical or state-of-the-art forgery detection methods, including Xception \cite{chollet2017xception}, TALL-Swin \cite{xu2023tall}, and BAR-Net \cite{zhang2024bandattention}.

Many models initially used for facial forgery detection, for example, three detection models in our benchmark, are general-purpose detection models, as they pay attention to the general differences introduced by generation instead of prior knowledge on facial regions.
These models are adaptable for body forgery detection as well, but they may lack guidance on prior knowledge regarding bodies. 
To further address the challenges in body forgery detection, we also propose \textbf{T}emporal \textbf{O}ptical \textbf{F}low \textbf{Net}work (TOF-Net).
Compared with human faces, human bodies exhibit a broader spatial range and more substantial magnitude of movements, thus incorporating motion information would be more advantageous for body forgery detection. 
TOF-Net integrates spatiotemporal attention and motion-guided optical flow modulation. It dynamically emphasizes moving areas by using the velocity norm of each pixel and focuses on pixels with abnormal color value changes, achieving favorable evaluation results on TT-DF.

\section{Related Work}
\subsubsection{Human Image Animation.} Human Image Animation involves driving the individuals in static images to generate human body videos based on the body poses and movements extracted from driving videos, where these pose signals are obtained through pose estimation \cite{cao2017realtime,simon2017hand,guler2018densepose,yang2023effective}. 
Earlier works, such as \cite{chan2019everybody,tulyakov2018mocogan}, primarily rely on GANs to achieve this process of motion transfer. 
Recent works mainly benefit from latent diffusion models. 
DreamPose \cite{karras2023dreampose} utilizes a diffusion model with image and pose conditions. 
DisCo \cite{wang2023disco} further emphasizes compositionality, which enables disentangled human foreground, background, and pose sequences to produce arbitrary compositions. 
MagicDance \cite{chang2023magicdance}, MagicAnimate \cite{xu2023magicanimate}, and AnimateAnyone \cite{hu2023animate} are three animation methods used for manipulating video data in TT-DF, which we will provide a more detailed introduction to in Section \ref{sec:manipulation}.

\subsubsection{Forgery Detection.} Forgery Detection aims to identify differences between natural and synthetic visual data that are imperceptible to the human eye, with facial forgery detection being a particularly crucial area within this field. 
Many early works \cite{afchar2018mesonet,nguyen2019capsule,rossler2019faceforensics++} rely on CNNs trained on cropped facial regions for feature extraction and classification, without fully exploiting facial prior knowledge, and often lead to overfitting.
Other approaches \cite{li2018ictu,yang2019exposing,huang2023implicit,bai2023aunet} focus on specific tracking and feature extraction for facial units or identities, lacking the extension to body forgery detection. 
Conversely, frequency-aware methods \cite{qian2020thinking,li2021frequency,zhang2024bandattention,luo2021generalizing,wang2023dynamic} represent a paradigm in focusing on the common characteristics of synthetic images, since previous works \cite{zhang2019detecting,frank2020leveraging} point out that upsampling in generative models can cause synthetic images to deviate from the natural frequency distribution.

\section{TT-DF: Human Body Forgery Dataset}

A core contribution of our work is the novel TT-DF dataset specifically tailored for body forgery. In this section, we concisely introduce the general principles of pose-driven human image animation and three approaches we select for manipulation: MagicDance \cite{chang2023magicdance}, MagicAnimate \cite{xu2023magicanimate}, and AnimateAnyone \cite{hu2023animate}.

\subsection{TT-DF: Manipulation Methods}\label{sec:manipulation}

\subsubsection{Latent Diffusion Models (LDMs).}

Given an image $I$, LDMs \cite{rombach2022high} utilize an autoencoder ${\mathcal{E}(\cdot)}$ \cite{van2017neural} to transform the input image to low-dimensional latent $z_0 = {\mathcal{E}}(I)$. During diffusion, $z_0$ is diffused in $T$ time steps to form $z_T \sim N(0,1)$. Conditional LDMs driven by text prompt $y$ accept $\tau_{\theta}(y)$ as the embedding prompt, and the learning objective is:

\begin{equation}
\mathcal{L} = \mathbb{E}_{{\mathcal{E}}(I), y, \epsilon \sim N(0,1), t} [
\left\| \epsilon - \epsilon_{\theta}(z_t, t, \tau_{\theta}(y)) \right\|^2_2
], t = 1, ..., T
\label{eq:condLDM}
\end{equation}
where prompt encoder $\tau_{\theta}$ and noise predictor $\epsilon_{\theta}$ are jointly optimized. During inference, $z_T \sim N(0,1)$ are denoised over $T$ steps to get $z_0$ and the generated image $\mathcal{D}(z_0)$ via Decoder $\mathcal{D}(\cdot)$.

LDM-based image animation models often require both pose signals and reference images, in which case \eqref{eq:condLDM} should be extended as:

\begin{equation}
\mathcal{L} = \mathbb{E}_{{\mathcal{E}}(I), r, p, \epsilon \sim N(0,1), t} [
\left\| \epsilon - \epsilon_{\theta}(z_t, t, \tau_{\theta}(r), \mu_{\theta}(p)) \right\|^2_2
], t = 1, ..., T
\label{eq:animationLDM}
\end{equation}
where $p$ and $r$ respectively denote pose condition and reference image prompt. Here, $\tau_{\theta}(r)$ typically serves as the image embedding prompt, fused with noisy latent via cross-attention, while a set of pose conditions $\mu_{\theta}(p)$ are often generated through downsampling and middle blocks in ControlNet \cite{zhang2023adding}. 

\subsubsection{MagicDance (MD).} MD \cite{chang2023magicdance} adopts OpenPose \cite{cao2017realtime} as its human pose detector. The training process of MD is divided into two stages: 

1) Appearance Control Pretraining. In this stage, a learnable Appearance Control Module is copied from SD (Stable Diffusion) -UNet and provides frozen SD-UNet with ID information through its Multi-Source Self Attention Module. The Appearance Control Model is trained with an objective similar to \eqref{eq:condLDM};

2) Appearance-disentangled Pose Control. In this stage, a pretrained Appearance Control Module is utilized to disentangle the pose ControlNet from ID information. Both Appearance Control Module and pretrained OpenPose ControlNet are involved in fine-tuning with an objective similar to \eqref{eq:animationLDM}.

\subsubsection{MagicAnimate (MA).} MA \cite{xu2023magicanimate} proposes its Appearance Encoder to replace the CLIP encoder preferred in earlier works \cite{karras2023dreampose,wang2023disco}. Explicit temporal attention blocks are also added to the original SD-UNet to mitigate inter-frame artifacts. Like MD, the training process of MA is also divided into two stages. In the first stage, both Appearance Encoder and pose ControlNet are trained with \eqref{eq:animationLDM}, while in the second stage, only additional temporal attention layers are optimized with \eqref{eq:animationLDM}, extended within continuously seen frames. MA also uses an image-video joint training strategy, with a probability threshold to decide whether to train on additional human image data from a large-scale image dataset \cite{schuhmann2021laion}.

OpenPose lacks sensitivity to rotation due to the sparseness of body keypoint representation, thus DensePose \cite{guler2018densepose} is utilized as a substitution in MA. In our implementation, we adopt a Detectron2 \cite{wu2019detectron2} DensePose model with Panoptic FPN head \cite{kirillov2019panoptic} and DeepLabV3 head \cite{chen2017rethinking} for pose estimation. 

However, unlike sparser skeleton representations, this denser pose format introduces additional ID information such as body shape and gender, so it does not completely decouple ID from pose sequence. As shown in \figref{fig1}, in MA's output of $Mismatch$ subset, ID from the reference image is somewhat distorted by ID from the pose image, characterized by a slight bulge in the male chest, while MD and AA exhibit no such phenomenon. Besides, DensePose is not adept at fine control, as it struggles to effectively control complex movements of fingers and facial expressions, which also limits the visual performance of MA. 

\subsubsection{AnimateAnyone (AA).} AA \cite{hu2023animate} uses a lightweight convolutional encoder, Pose Guider, instead of ControlNet, whose parameter count aligns with that of SD-UNet, and pose embedding is directly added into the noise latent with the same resolution. AA also incorporates temporal attention layers into denoising UNet. It introduces a temporal-free counterpart, ReferenceNet, for reference image feature extraction, complemented by a CLIP encoder for prompts. 

For TT-DF, we use the $Moore-AnimateAnyone$ unofficial implement \cite{moore-AnimateAnyone}. 

\begin{figure} \centering \includegraphics[width=1\textwidth]{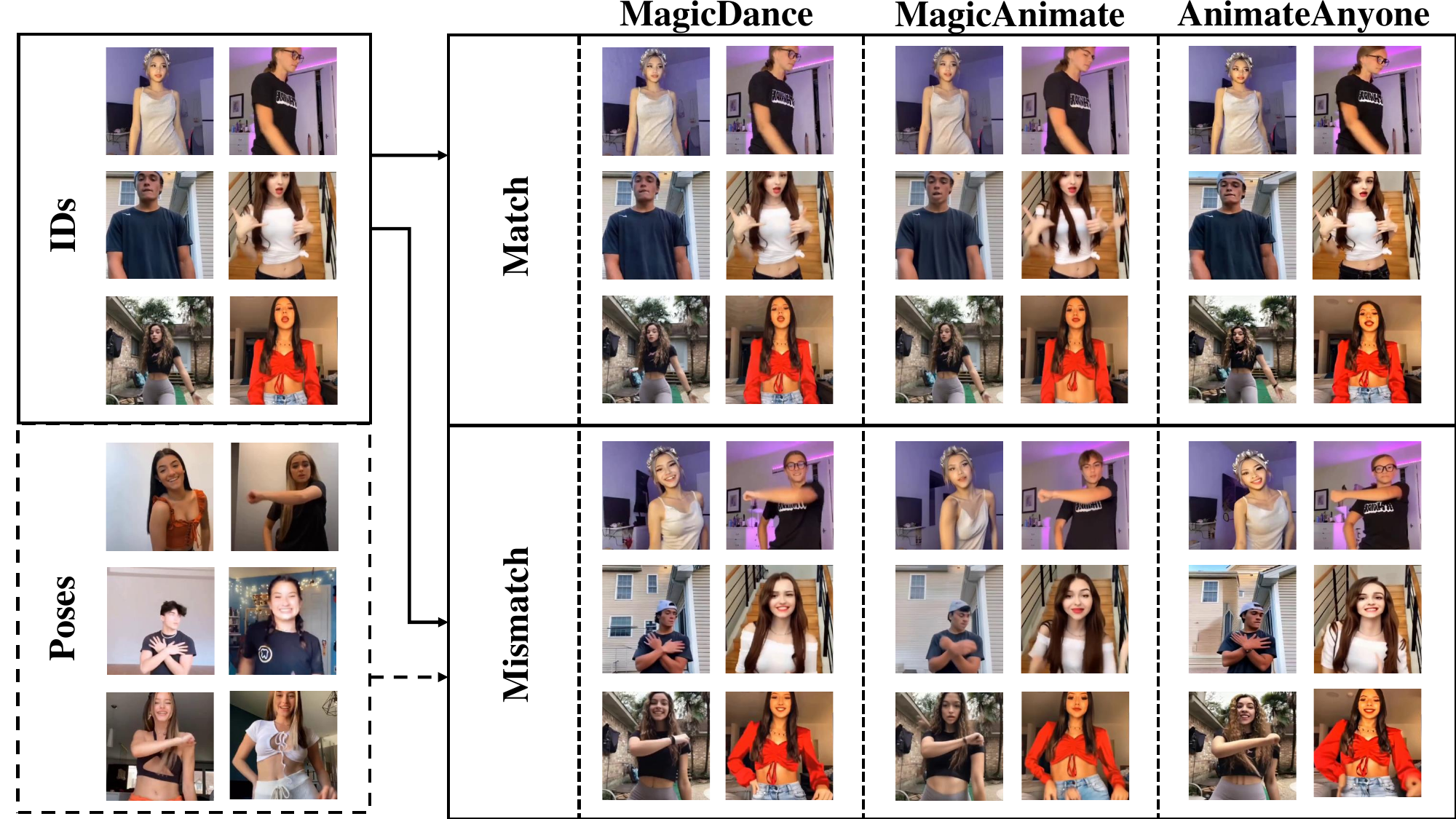} \caption{More examples in TT-DF. Due to the reliance on DensePose, MagicAnimate lacks precise control over fingers and expressions, and there is also leakage of pose ID.}
\label{fig2} \end{figure}

\subsection{TT-DF: Dataset Structure}
To ensure consistency between training and inference, we use one of their shared training datasets, TikTok \cite{jafarian2021learning}, for forged data generation. TikTok comprises 340 videos capturing single-person dance performances, with a frame rate of 30FPS. 

For one single original video in TikTok, we 1) utilize the first frame from the original TikTok video as the \textbf{R}eference \textbf{I}mage (RI), representing ID information; and 2) extract the \textbf{P}ose \textbf{S}equence (PS) via pose estimation. Then $\textbf{Match}$ and $\textbf{Mismatch}$ subsets, with distinct generative configurations, are generated according to whether RI and PS are from the same video. For example, $RI1$ and $PS1$ both from $Video1$ are combined to create a $Match$ forged video, while $RI1$ from $Video1$ and $PS2$ from $Video2$ are combined to create a $Mismatch$ forged one, as shown in \figref{fig1} and \figref{fig2}.

\textbf{Match} subsets are generated for high-quality video data because the three aforementioned animation methods are not only trained on TikTok but also evaluated quantitatively on TikTok in terms of reconstruction metrics. 
Given this realistic scenario, we perform $Match$ generation, hoping to obtain the highest possible quality of video data. 
On the other hand, \textbf{Mismatch} subsets are generated for video data closer to the actual forgery distribution. Considering that the user-customized demand in reality often involves transferring one pose sequence to another person's identity, we conduct $Mismatch$ generation with mismatched RI and PS. Specifically, we divide 340 videos into 170 pairs and exchange RI and PS in the same pair. 

In practice, forged videos also undergo lossy compression through multiple rounds of dissemination on social platforms, thereby compromising the artifacts relied upon by human detection.
Taking this into account, we compress all the videos in TT-DF using the H.264 codec with Constant Rate Factor (CRF) set to 23 and 40, consistent with FaceForensics++ (FF++) \cite{rossler2019faceforensics++}.

\section{Human Body Forgery Detection}
\begin{figure}
\centering
\includegraphics[width=0.95\textwidth]{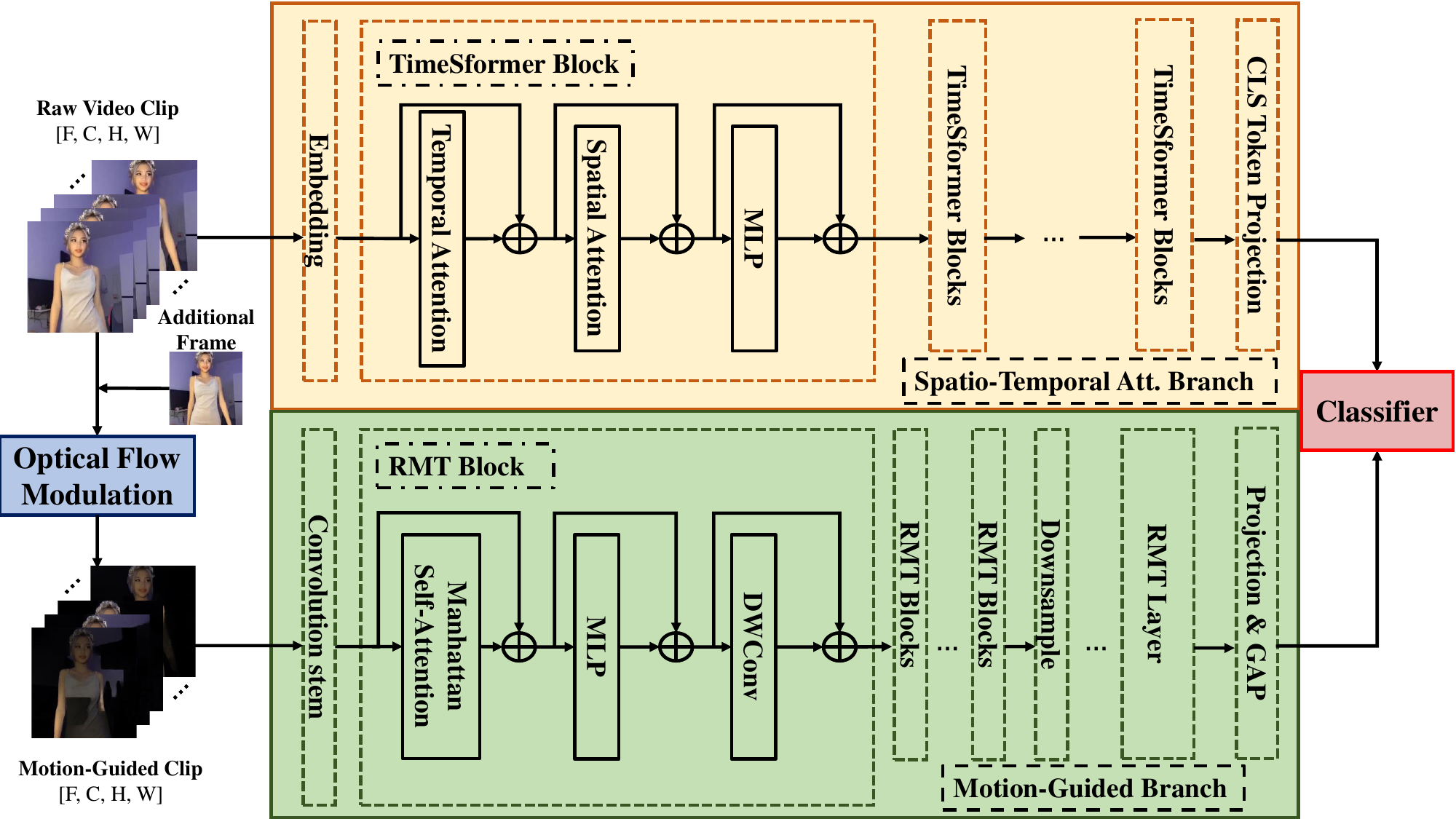}
\caption{Overview of \textbf{T}emporal \textbf{O}ptical \textbf{F}low \textbf{Net} (TOF-Net). TOF-Net comprises two branches: the Spatio-Temporal Attention branch and the Motion-Guided branch, which are designed to extract features of pose sequences and spatiotemporal ID information respectively. For better visualization of Motion-Guided Clip, the calculation method for Optical Flow Modulation here differs slightly from \eqref{eq:OFM}.} \label{fig3}
\end{figure}

Another core contribution of this work is our body forgery detection model, \textbf{T}emporal \textbf{O}ptical \textbf{F}low \textbf{Net} (TOF-Net), which is illustrated in \figref{fig3}.

The spatial range and amplitude of human postural movements are significantly greater compared to facial regions. Therefore, the integration of motion information has greater potential for body forgery detection. Considering the interaction between ID and pose information, TOF-Net adopts two relatively independent branches. It utilizes spatiotemporal attention to focus on ID information, and extracts feature frame-by-frame on motion-guided frames obtained by Optical Flow Modulation to get pose information.

\subsubsection{Spatio-Temporal Attention Branch.}
We sample consecutive $F\ (=8)$ frames from one video, obtaining clips with the shape of $\mathbb{R}^{F \times C \times H \times W}$. Subsequently, this video clip is fed into TimeSformer \cite{bertasius2021space} pretrained on Kinetics-600 \cite{carreira2018short}, utilizing Divided Space-Time Attention variation. For inter-frame temporal attention, we apply a self-attention mechanism to patches of the same spatial position across frames. For intra-frame spatial attention, the classic patch-based self-attention used in vanilla ViT \cite{dosovitskiy2020vit} is employed. Through this backbone, the global features obtained via intra-/inter-frame self-attention within a video clip are condensed into one classification token, which participates in the final detection.

\subsubsection{Motion-Guided Branch.}
Many animation models \cite{wu2023tune,guo2023animatediff,xu2023magicanimate,hu2023animate}, including MA and AA, incorporate temporal attention in their UNet. However, despite extending the two-dimensional per-frame MSE loss to a per-clip MSE loss, the pretrained SD performs on the image level rather than the video level. 
At best, this extension achieves isotropic improvement across up to three dimensions, and may even be constrained by the training volume of fine-tuning methods, unable to guarantee this isotropy. The qualitative results of TT-DF also indicate that the current shortcomings of animation methods lie in inter-frame artifacts. Based on these facts and analysis, we argue that the visible artifacts of concern in video forgery detection predominantly center around temporal inconsistencies between frames.

This temporal inconsistency is essentially abnormal changes in the pixel color values, which further originates from the dissonance of motion. Therefore, we can filter out this abnormal color value change by optical flow estimation. We use optical flow modulation (OFM) that operates on color values in pixel space to distinguish moving areas from static areas. As shown in \figref{fig3}, this area of interest is concentrated near human bodies, corresponding to the pose modality.

Specifically, we utilize a frozen RAFT module \cite{teed2020raft} for optical flow estimation. Then we compute the norms of vectors as their velocity values $|| \vec{\nu} ||$ for each pixel based on their velocity vector $\vec{\nu} = (u, v)$, where $u$ and $v$ respectively represent the horizontal and vertical velocity components. Subsequently, we perform normalization within the frame to obtain a weight map based on optical flow, which is then combined with the original frame as:

\begin{equation}
\mathrm{OFM}(frame) = (frame \odot \mathrm{RAFT}(frame) + frame) / 2
\label{eq:OFM}
\end{equation}

where $\odot$ denotes the Hadamard product. To calculate the optical flow of the last frame, an additional frame needs to be introduced at the end. Motion-guided frames are then delivered to RMT backbone \cite{fan2023rmt} for feature extraction. In contrast to the spatiotemporal branch, the motion-guided branch involves no interaction between frames in its backbone. Thus, the output of this branch is frame-by-frame features, which are later concentrated together before being sent to the final projection layer for classification.

\section{Experiment}
\subsection{Experimental Setup}
\subsubsection{Dataset.}
The resolution of all 340 videos in the original TikTok dataset is standardized to 604×1080. To ensure uniform preprocessing, we center-crop all the generated fake data and original data from TikTok to 604×604 and then scale them to a resolution of 512×512 in TT-DF.

In the 340 original videos (170 pairs), we select 240 (120 pairs) for the training set, 50 (25 pairs) for the validation set, and 50 (25 pairs) for the test set. The forged videos in the $Match$ subset are individually assigned to the train/val/test set based on the same partition as the original videos. For the forged videos in the $Mismatch$ subset, they are paired and assigned to these sets based on the same partition as the original video pairs. There are 200 test videos in $Match$ and $Mismatch$ subsets. We clip all test videos to segments of fewer than 30 frames, resulting in 1,353 $Mismatch$ and 1,519 $Match$ test video clips, to mitigate fluctuations in the test metrics.

\subsubsection{Implementation Details.}
We evaluate our method on the proposed TT-DF dataset, and we also provide three benchmarks on it, including classical Xception \cite{chollet2017xception} which can be found in many facial forgery benchmarks, and two additional methods: TALL-Swin \cite{xu2023tall} and BAR-Net \cite{zhang2024bandattention}. Among them, Xception and BAR-Net are image forgery detection methods, while TALL-Swin is a video forgery detection method. We extend Xception and BAR-Net to a video detection model by predicting frame-wise and averaging the results between frames. All three models undergo the same data preprocessing steps. For evaluation, we utilize Accuracy (Acc) and Area Under the Receiver Operating Characteristic Curve (AUC) metrics, consistent with most prior research on facial forgery detection.

TOF-Net accepts video clips with a fixed number $(F = 8)$ of frames as its inputs. To ensure the effectiveness of the additional frames required for optical flow estimation, we sample consecutive $(F + 1)$ frames in a video during training. The starting frame is pseudo-randomly selected within the range $[0,\ len(video) - F - 1]$ to ensure sufficient feature learning through multiple pseudo-random samplings for each training video. During evaluation and testing, to obtain the most stable and credible prediction results, we uniformly sample the starting frames multiple times within the same range, without randomness.

\subsection{Intra-dataset Evaluation}

In this section, we first compare our model with other benchmark models within the Match subsets of TTDF-C23/C40, which represent the aforementioned H.264 compressed versions with CRF set to 23/40. We train all these models on the TTDF-C23/C40 \textbf{Match} subsets and test them on the same subsets.

\begin{table}
\renewcommand{\arraystretch}{1.3}
\setlength{\tabcolsep}{4pt}
\centering
\caption{Comparison of video-level test results within the TT-DF $\textbf{Match}$ subsets. We train on the higher-quality $\textbf{Match}$ subsets and test on the $\textbf{Match}$ subsets. Here TTDF-C23/C40 denote compressed version with CRF set to 23/40. The top two rankings are highlighted in \textcolor{red}{red} and \textcolor{blue}{blue}, respectively.}\label{tab1}
\begin{tabular}{@{}lcccc@{}}
\toprule
\multirow{2}{*}{Method} &  \multicolumn{2}{c}{TTDF-C40} & \multicolumn{2}{c}{TTDF-C23} \\ 
\cmidrule(lr){2-3} \cmidrule(lr){4-5} & AUC (\%)  & Acc (\%) & AUC (\%) & Acc (\%) \\ \midrule
Xception \cite{chollet2017xception} & 90.95 & 87.82 & \color{blue} 99.17 & 96.25 \\
TALL-Swin \cite{xu2023tall} & 91.88 & 87.01 & 95.57 & 90.44 \\
BAR-Net \cite{zhang2024bandattention} & \color{blue} 92.92 & \color{blue} 88.74 & \color{red} 99.65 & \color{blue} 96.91 \\ \hline
\textbf{Ours} & \bf \color{red} 94.76 & \bf \color{red} 89.99 & \bf 99.11  & \bf \color{red} 97.04  \\ \bottomrule
\end{tabular}
\end{table}

As depicted in Tab. \ref{tab1}, our proposed method attains the highest performance on TTDF-C40 with an AUC of 94.76\%, surpassing the second-ranked BAR-Net with 92.92\%, and it also achieves the highest Acc results on both C23 and C40. Despite C40 exhibiting lower visual quality compared to C23, the comparison results of TOF-Net with other models on C40 still outperform those on C23. This can be attributed to the motion-guided approach directing the model's attention more towards coarse-grained motion inconsistencies rather than detailed textures and artifacts hidden in high-frequency components. Additionally, this suggests that our method is not dependent on more expensive high-quality video data.

\subsection{Generalization Ability Evaluation}
For forgery detection models, the importance of their generalization ability far outweighs their performance within the datasets. These models, once trained, need to face the challenge of unseen forgery methods, which may involve adopting new configurations on known models or using entirely new models for manipulation. Therefore, in this section, we evaluate the generalization ability of these benchmark detection models from these two perspectives.

\subsubsection{Cross-configuration evaluation (CCE).}
We perform CCE to assess how benchmark models perform under different generation configurations. Specifically, we train the models on the TTDF-C23/C40 \textbf{Match} subsets. However, unlike the previous section, we test these trained models on the \textbf{Mismatch} subsets generated with a different configuration. 

\begin{table}
\renewcommand{\arraystretch}{1.3}
\setlength{\tabcolsep}{4pt}
\centering
\caption{Comparison of video-level test results within the TT-DF $\textbf{Mismatch}$ subsets. We train on the higher-quality $\textbf{Match}$ subsets but test on the $\textbf{Mismatch}$ subsets, which are closer to the real distribution. The top two rankings are highlighted in \textcolor{red}{red} and \textcolor{blue}{blue}, respectively.} \label{tab2}
\begin{tabular}{@{}lcccc@{}}
\toprule
\multirow{2}{*}{Method} &  \multicolumn{2}{c}{TTDF-C40} & \multicolumn{2}{c}{TTDF-C23} \\ 
\cmidrule(lr){2-3} \cmidrule(lr){4-5} & AUC (\%)  & Acc (\%) & AUC (\%) & Acc (\%) \\ \midrule
Xception \cite{chollet2017xception} & 89.74 & 86.40 & 98.60 & \color{blue} 95.56 \\
TALL-Swin \cite{xu2023tall} & \color{blue} 91.91 & 86.46 & 95.83 & 90.46 \\
BAR-Net \cite{zhang2024bandattention} & 91.07 & \color{blue} 87.43 & \color{blue} 98.68 & \color{red} 96.16 \\ \hline
\textbf{Ours} & \bf \color{red} 93.82 & \bf \color{red} 87.58 & \bf \color{red} 98.76  & \bf 95.34  \\ \bottomrule
\end{tabular}
\end{table}

As depicted in Tab. \ref{tab2}, in CCE, our model still performs well on C40, achieving the highest AUC on both C23 (98.76\%) and C40 (93.82\%). We observe a decrease in overall metrics compared to the intra-dataset evaluations on the Match subsets (Tab. \ref{tab1}). On one hand, the quality of videos in the \textbf{Mismatch} subsets, both visually and quantitatively, is inferior to that of the \textbf{Match} subsets, which reduces the difficulty of detection. On the other hand, there do exist distributional gaps between these two kinds of videos, which conversely increases the difficulty. According to our experiments, we can conclude that the decrease in metrics is primarily caused by the latter factor.

\subsubsection{Cross-manipulation evaluation (CME).}
The field of human body generation is still rapidly evolving, and correspondingly, good detection models should exhibit excellent performance while facing unseen manipulation models. To simulate this scenario, similar to FF++, we conduct CME within TTDF-C40 $Match$. Specifically, we train on training sets generated by two manipulation models and real data, and test on the test set generated by the remaining model along with real data. These results are presented in Tab. \ref{tab3}, where CME-2MD indicates training on MA and AA and testing on MD, and so forth.

\begin{table}
\renewcommand{\arraystretch}{1.3}
\setlength{\tabcolsep}{4pt}
\centering
\caption{Comparison of video-level test results within TTDF-C40 $\textbf{Match}$. We train on the training sets of two forgery models and the real data, and then test on the test sets of the remaining model and the real data. Here CME-2MD indicates training on MA and AA and testing on MD, and so forth. The top two rankings are highlighted in \textcolor{red}{red} and \textcolor{blue}{blue}, respectively.} \label{tab3}
\begin{tabular}{@{}lcccccc@{}}
\toprule
\multirow{2}{*}{Method} &  \multicolumn{2}{c}{CME-2MD} & \multicolumn{2}{c}{CME-2MA} & \multicolumn{2}{c}{CME-2AA} \\ 
\cmidrule(lr){2-3} \cmidrule(lr){4-5} \cmidrule(lr){6-7} & AUC (\%)  & Acc (\%) & AUC (\%) & Acc (\%)  & AUC (\%) & Acc (\%) \\ \midrule
Xception \cite{chollet2017xception} & 75.58 & 69.88 & 68.23 & 65.78 & \color{blue} 88.30 & \color{blue} 81.24 \\
TALL-Swin \cite{xu2023tall} & \color{blue} 77.58 & \color{red} 72.58 & 70.42 & 65.86 & 84.16 & 76.14 \\
BAR-Net \cite{zhang2024bandattention} & 76.48  & 68.67 & \color{blue} 73.55 & \color{red} 68.95 & 85.14 & 77.44 \\ \hline
\textbf{Ours} & \bf \color{red} 79.98  & \bf \color{blue} 72.16  & \bf \color{red} 74.39 & \bf \color{blue} 68.57 & \bf \color{red} 90.04 & \bf \color{red} 81.50 \\ \bottomrule
\end{tabular}
\end{table}

For all models, the performance on CME-2MA is the poorest, which is attributed to the distributional differences caused by the distinct pose estimation methods employed in MA. Following this, CME-2MD exhibits the next level of performance, but MD uses per-frame MSE rather than per-clip MSE as its loss. CME-2AA achieves the best results, as both its pose estimation and training objectives appear in the training set. Our model achieves the best AUC results in all three CME tests, indicating that motion-aware prior knowledge can uncover inter-frame inconsistencies beyond specific manipulation models. 

\subsection{Ablation Study}

TOF-Net comprises two branches: the spatio-temporal attention branch (S-T branch) and the motion-guided branch (M-G branch). The S-T branch obtains classification tokens for the entire clips, while the M-G branch concatenates per-frame features, which are later fed into an MLP to balance its parameter count with the S-T branch. Together, they undergo final binary classification. To assess the individual contributions of each branch, we conduct an ablation study as shown in Tab. \ref{tab4}. Overall, the performance of the M-G branch alone is superior to that of the S-T branch.

\begin{table}
\renewcommand{\arraystretch}{1.3}
\setlength{\tabcolsep}{4pt}
\centering
\caption{Comparison of ablation study results within the TT-DF $\textbf{Match}$ subsets. We train on the $\textbf{Match}$ subsets and test on the $\textbf{Match}$ subsets. Here $S-T \ branch$ denotes Spatio-Temporal attention branch, while $M-G \ branch$ denotes Motion-Guided branch.} \label{tab4}
\begin{tabular}{@{}lcccc@{}}
\toprule
\multirow{2}{*}{Method} &  \multicolumn{2}{c}{TTDF-C40} & \multicolumn{2}{c}{TTDF-C23} \\ 
\cmidrule(lr){2-3} \cmidrule(lr){4-5} & AUC (\%)  & Acc (\%) & AUC (\%) & Acc (\%) \\ \midrule
S-T branch & 93.51 & 87.29 & 95.54 & 91.19 \\
M-G branch & 93.63 & 88.02 & 97.64 & 92.96 \\ \hline
\textbf{Ours} & \bf 94.76 & \bf 89.99 & \bf 99.11  & \bf 97.04 \\ \bottomrule
\end{tabular}
\end{table}

\section{Conclusion}
In this paper, we introduce TT-DF, a novel large-scale diffusion-based dataset tailored for human body forgery detection. To the best of our knowledge, TT-DF is the pioneer dataset dedicated to this emerging area, encompassing various compressed versions, advanced diffusion-based animation models, and generation configurations. We also propose Temporal Optical Flow Network (TOF-Net) for body forgery detection, comprising a motion-guided branch with optical flow estimation and a spatio-temporal attention branch. Additionally, we evaluate several forgery detection methods and our proposed TOF-Net, establishing a benchmark on TT-DF. We hope that our TT-DF dataset and benchmark will catalyze advancements in forgery detection and enhance AI security.

\subsubsection{Acknowledgement.}
This work is funded by Beijing Municipal Science and Technology Project (Nos. Z231100010323005), National Natural Science Foundation of China (Grant No. 62206277), Beijing Nova Program (20230484276), and Youth Innovation Promotion Association CAS (Grant No. 2022132).

%
%
%
\bibliographystyle{splncs04}
\bibliography{main}

\end{document}